%% file: acl_latex.tex
\pdfoutput=1
\documentclass[11pt]{article}

\usepackage[preprint]{acl}

\usepackage{times}
\usepackage{latexsym}

\usepackage[T1]{fontenc}
\usepackage[utf8]{inputenc}

\usepackage{microtype}

\usepackage{float}
\usepackage{newfloat}
\usepackage{multirow}
\usepackage{makecell}
\usepackage{xcolor}  
\usepackage{colortbl}
\usepackage{dsfont}
\usepackage{amsmath}
\usepackage{enumitem}
\usepackage{todonotes}
\usepackage[normalem]{ulem}
\usepackage{fontawesome}
\usepackage{amssymb}
\usepackage{booktabs}
\usepackage{stfloats}
\usepackage{tabularx}

\useunder{\uline}{\ul}{}

\definecolor{cvprblue}{rgb}{0.21,0.49,0.74}
\definecolor{myyellow}{rgb}{1,1, 0.6}
\definecolor{myorange}{rgb}{1, 0.8, 0.6}
\definecolor{myred}{rgb}{1, 0.6, 0.6}
\definecolor{lightblue}{rgb}{0.8,0.9,1}
\definecolor{second}{HTML}{FFDAB9}
\definecolor{best}{HTML}{FFC1C1}
\newcommand*{\ourmethod}{IDEALPrompt}
\newcommand*{\ourstagei}{Reinforcement Warm-up Strategy}
\newcommand*{\ourstageii}{Empirical Self-reflective Optimization}
\newcommand*{\ourstageishort}{RWS}
\newcommand*{\ourstageiishort}{ESO}
\newcommand*{\benchmark}{\texttt{Taobao-PDA}}

\renewcommand{\todo}[1]{\iffalse #1 \fi{\color{blue} \textbf{[TODO]}}}
\usepackage{hyperref}

\title{Boosting Private Domain Understanding of Efficient MLLMs: A Tuning-free, Adaptive, Universal Prompt Optimization Framework}

\author {
    \small{
    \textbf{Jiang Liu\textsuperscript{\rm 1, \rm \dag}},
    \textbf{Bolin Li\textsuperscript{\rm 2, \rm \dag}},
    \textbf{Haoyuan Li\textsuperscript{\rm 2, \rm \dag}},
    \textbf{Tianwei Lin\textsuperscript{\rm 1}},
    \textbf{Wenqiao Zhang\textsuperscript{\rm 1, \rm *}},
    \textbf{Tao Zhong\textsuperscript{\rm 2}},
    \textbf{Zhelun Yu\textsuperscript{\rm 2}},
    \textbf{Jinghao Wei\textsuperscript{\rm 2}},
    } \\
    \small{
    \textbf{Hao Cheng\textsuperscript{\rm 2}},
    \textbf{Wanggui He\textsuperscript{\rm 2}},
    \textbf{Fangxun Shu\textsuperscript{\rm 2}},
    \textbf{Hao Jiang\textsuperscript{\rm 2, \rm *}},
    \textbf{Zheqi Lv\textsuperscript{\rm 1, \rm *}},
    \textbf{Juncheng Li\textsuperscript{\rm 1}},
    \textbf{Siliang Tang\textsuperscript{\rm 1}},
    \textbf{Yueting Zhuang\textsuperscript{\rm 1}}
    }
\\
    \small \textsuperscript{\rm 1}Zhejiang University\\
    \small \textsuperscript{\rm 2}Alibaba Group\\
    \small \{jiang.liu, lintw, wenqiaozhang, zheqilv, junchengli, siliang, yzhuang\}@zju.edu.cn;\\
    \small \{libolin.lbl, lihaoyuan.lhy, zt395565, yuzhelun.yzl, weijinghao.wjh, chenghao.x, wanggui.hwg, shufangxun.sfx, aoshu.jh\}@taobao.com
}

\begin{document}
\maketitle

\input{section/0_abstract}

\input{section/1_introduction}
\input{section/2_related_work}
\input{section/3_method}
\input{section/4_experiment}

\input{section/5_conclusion}

\clearpage
\input{section/6_limitation}

\bibliography{custom}

\clearpage
\appendix
\input{section/X_suppl}

\end{document}

%% file: section/0_abstract.tex
\begin{abstract}

% Efficient multimodal large language models (EMLLMs), compared to multimodal large language models (MLLMs), reduce model size and computational costs and are often deployed on resource-constrained devices. 
Efficient multimodal large language models (EMLLMs), in contrast to multimodal large language models (MLLMs), reduce model size and computational costs and are often deployed on resource-constrained devices.
However, due to data privacy concerns, existing open-source EMLLMs rarely have access to private domain-specific data during the pre-training process, making them difficult to directly apply in device-specific domains, such as certain business scenarios. 
% This manifests as an inability to understand specific prompts in these scenarios. Additionally, fine-tuning EMLLMs on all private data for domain-adaptive learning requires substantial resources, and the high training costs significantly raise the application threshold, severely limiting their availability and generalizability on devices. 
To address this weakness, this paper focuses on the efficient adaptation of EMLLMs to private domains, specifically in two areas: 1) how to reduce data requirements, and 2) how to avoid parameter fine-tuning. 
Specifically, we propose a tun\textbf{\underline{I}}ng-free, a\textbf{\underline{D}}aptiv\textbf{\underline{E}}, univers\textbf{\underline{AL}} \textbf{\underline{Prompt}} Optimization Framework, abbreviated as \textit{\textbf{\ourmethod{}}} which consists of two stages: 1) Predefined Prompt, based on the reinforcement searching strategy, generate a prompt optimization strategy tree to acquire optimization priors; 2) Prompt Reflection initializes the prompt based on optimization priors, followed by self-reflection to further search and refine the prompt. By doing so, \ourmethod{} elegantly generates the ``ideal prompts'' for processing private domain-specific data. Note that our method requires no parameter fine-tuning and only a small amount of data to quickly adapt to the data distribution of private data.
Extensive experiments across multiple tasks demonstrate that our proposed \ourmethod{} significantly improves both efficiency and performance compared to baselines.
% ~\footnote{Our project is available at \url{https://anonymous.4open.science/r/IDEALPrompt-01DB}.}

\end{abstract}

%% file: section/1_introduction.tex
\section{Introduction}
\label{sec:intro}

In recent years, multimodal large language models (MLLMs) have achieved significant advancements, particularly in cross-modal information understanding and integration \citep{huang2023language,ref:MM-GPT,ref:mPLUG-Owl,li2023traineragent}.
Efficient multimodal large language models (EMLLMs) significantly reduce model size and computational costs compared to MLLMs, making them ideal for deployment on resource-constrained devices for enhanced processing efficiency \citep{jin2024efficientmultimodallargelanguage}.
However, the scenarios encountered by different devices, including various private business contexts, can vary significantly.
Moreover, ethical considerations and the necessity to protect trade secrets prevent the use of device data for pre-training by open-source, general-purpose EMLLMs.
Consequently, a discrepancy between the distribution of pre-training data and device-specific data emerges, leading to reduced performance or even failure when EMLLMs process this unique and highly private domain data.

A viable solution involves utilizing EMLLMs, \emph{e.g.}, InternVL2-2B \citep{chen2023internvl, chen2024far} and Qwen2-VL-2B \citep{Qwen2VL}, to perform either full-parameter \citep{chung2024scaling, zhang2024hyperllava} or efficient fine-tuning \citep{hu2021lora, lin2024teamlora} on private device data for domain-specific adaptation \citep{chang2019domain, zhang2022boostmis, zhang2024revisiting, Li_2023_CVPR}.
However, this approach incurs large training costs, necessitates large-memory GPUs, and entails long training times.
The elevated training costs significantly increase the application threshold, thereby restricting the availability and generalizability of such methods on private domain data. 
% Furthermore, while the reflective CoT approach is capable of addressing complex problems, it struggles to comprehend private-domain data due to the data distribution gap between private and public-domain data.
Therefore, our goal is to minimize the adaptation cost for EMLLMs when deployed on devices utilizing private domain-specific data.

\begin{figure*}
    \centering
    \includegraphics[width=0.99\linewidth]{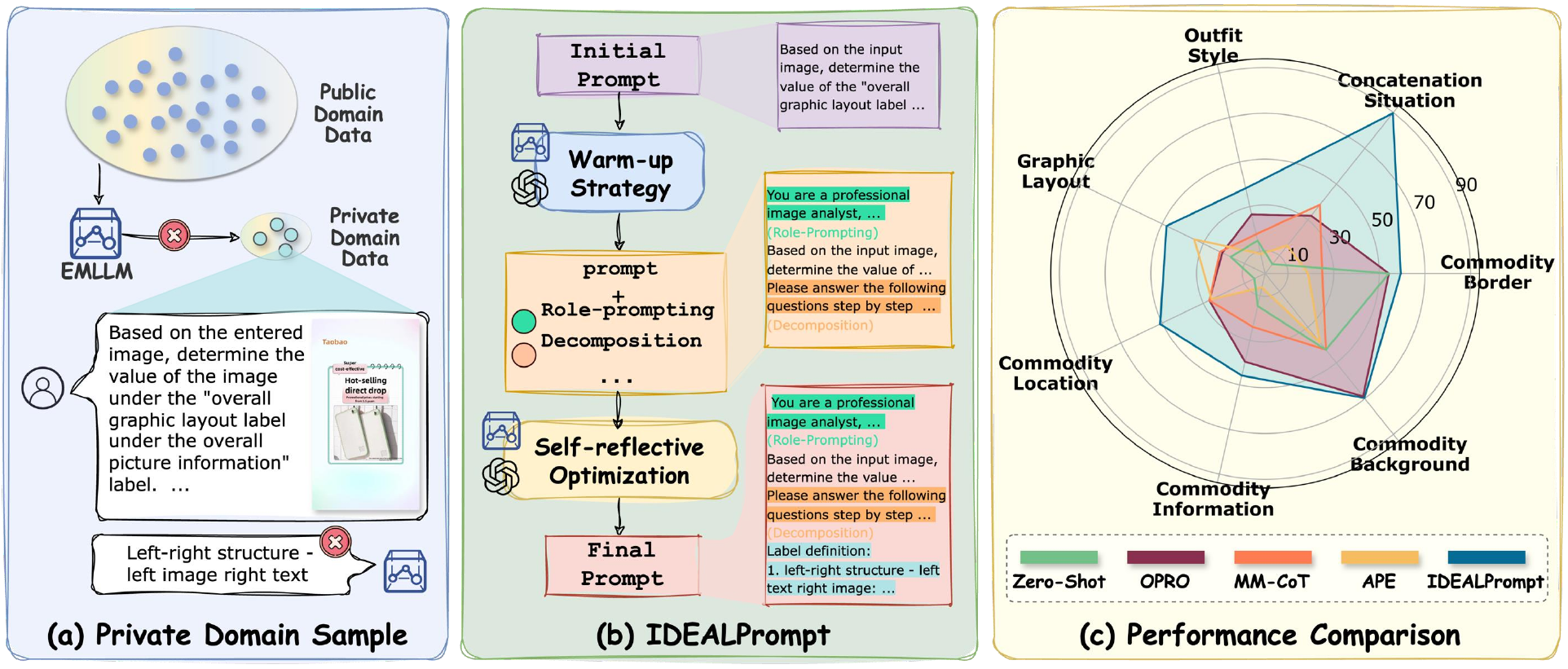}
    \caption{(a) shows the gap between public domain data and private domain data. (b) describes the simplified version of our \ourmethod{}. (c) illustrates that compared to baseline methods, our method achieves superior performance on \benchmark{} benchmark.}
    \label{fig:front}
    \vspace{-0.5cm}
\end{figure*}

In response to the aforementioned challenges, we propose a tun\textbf{\underline{I}}ng-free, a\textbf{\underline{D}}aptiv\textbf{\underline{E}}, univers\textbf{\underline{AL}} \textbf{\underline{Prompt}} optimization framework, referred to as \textbf{\ourmethod{}}.
Our bootstrapping philosophy aims to strengthen EMLLMs' capability in private domain tasks by progressively and adaptively optimizing the prompt.
% utilizing a limited number of annotated training samples and the occasional use of capable MLLMs for prompt optimization.
% Our approach uses a reinforcement learning approach that allows for prompt optimization without requiring extensive training data or costly iterative tuning. The two-stage framework leverages a limited number of annotated training samples and occasional use of capable MLLMs for prompt optimization, enabling EMLLMs to rapidly enhance their capability in private domain tasks.
\ourmethod{} consists of two progressive optimization steps: \textbf{i) \ourstagei{} (\ourstageishort{})}, which employs a reinforcement learning tree search strategy for the general prompt optimization, encouraging combinations of diverse prompt strategies that are most effective given the context of the current task.
% effective prompt optimization strategy tree search based on Reinforcement Learning(RL),
% allowing the framework to explore combinations of prompt strategies that are most effective for given the context of the current task.
This step enables the framework to gather prior knowledge that can be adapted across tasks and models, significantly reducing iteration cycles while maintaining performance.
It effectively functions as a ``Pre-Training'' phase for task-specific prompt engineering, similar to the pre-training in the MLLMs' training paradigm, thereby encouraging the framework's adaptability to varied task requirements.
\textbf{ii) \ourstageii{} (\ourstageiishort{})}, further refines the prompts by selecting critical bad cases from the error distribution of inference results and incorporating self-reflection. 
By focusing on these critical bad cases, \ourmethod{} learns to optimize prompts by identifying and correcting ambiguous task descriptions and labels. This step allows the framework to continuously improve its understanding of prompt definitions in private domain tasks, functioning as a ``Supervised Fine-Tuning'' phase for prompt refining, similar to the supervised fine-tuning in the MLLMs' training paradigm, thereby boosting the framework's performance.

To evaluate our method in private domain adaptation thoroughly, we introduce the Taobao Private Domain Adaptation (\benchmark{}), a new Chinese benchmark comprising multiple complex multimodal tasks, which is able to evaluate the private domain understanding capability of EMLLMs.
% \ourmethod{} achieves significant results on the proposed Taobao private domain multi-task multimodal content understanding (\benchmark{}) benchmark, demonstrating that it can efficiently boost model performance in complex, multi-task settings. 
The results of the experiment demonstrate that \ourmethod{} exhibits strong generalization and adaptability across multiple tasks, achieving robust performance in content understanding while minimizing computational costs. Our contributions are as follows:
\begin{itemize}[leftmargin=1em, itemsep=0em, parsep=0em, topsep=0em]
    \item To the best of our knowledge, this is the first study that focuses on the efficient adaptation of on-device EMLLMs to private data distributions that differ significantly from the pre-trained data distribution.
    \item We meticulously devise a pre-made prompt dataset for multimodal business scenarios, called \benchmark{}, which serves as fundamental data support for promoting private domain understanding.
    % which, to the best of our knowledge, is the first of its kind.
    % It can be widely applied to EMLLMs and MLLMs.
    \item We propose \ourmethod{}, a tuning-free, adaptive, universal prompt optimization framework, which leverages a two-stage design to facilitate EMLLMs' content understanding of private domain data.
    % \ourstagei{} and \ourstageii{}. The former adapts to the data distribution of device-specific private data solely through prompt-based methods, thereby reducing adaptation costs by avoiding parameter fine-tuning of EMLLMs. The latter selects the most effective data for model adaptation by filtering out bad cases, which reduces the data volume required for adaptation and further lowers adaptation costs.
    \item Extensive and comprehensive experiments are conducted and followed by detailed analysis, which demonstrates the superiority of our method compared to the baselines. \textbf{Notably, our method even outperforms the fine-tuning method on proposed \benchmark{}.}
\end{itemize}

%% file: section/2_related_work.tex
\section{Related Work}
\label{sec:related_work}

% 单模态Prompt Optimization会不会好点
\noindent \textbf{Prompt Optimization.}
Prompt optimization can enhance the performance of MLLMs without parameter fine-tuning, and several methods have been proposed to address the challenges of automated prompt optimization. Recent works focus on discrete optimization methods, such as GRIPS \citep{prasad2022grips} and APO \citep{pryzant2023automatic}, which utilize edit-based operations to generate various candidate prompts for subsequent optimization. Other approaches leverage advanced algorithms for prompt optimization, including EvoPrompt \citep{guo2023connecting}, PromptBreeder \citep{fernando2023promptbreeder}, AELP \citep{hsieh2023automatic}, and PhaseEvo \citep{cui2024phaseevo}, which leverage evolutionary algorithms to produce different prompts; RLPROMPT \citep{deng2022rlprompt} and PRewrite \citep{kong2024prewrite} apply reinforcement learning techniques; PromptAgent \citep{wang2023promptagent} and Agent-Pro \citep{zhang2024agent} utilize agent-based methods. APE \citep{zhou2022large} generates candidate prompts using MLLMs, followed by a resampling of those prompts. OPRO \citep{Yang2023LargeLM} leverage MLLMs as prompt optimizers, iteratively generating superior prompts by considering previously generated solutions with their values. Additionally, some research reformulates prompt optimization as a continuous optimization problem, as seen in InstructZero \citep{chen2023instructzero}, INSTINCT \citep{pmlr-v235-lin24r}, and ZOPO \citep{hu2024localized}. 
APOHF \citep{lin2024prompt} prompt optimization based on human preference feedback rather than specific numerical scores. IPC \citep{levi2024intent} aims to optimize prompt engineering for MLLMs by utilizing synthetic case samples. 
% Furthermore, prompt optimization can also be applied to address other tasks, such as in prompt optimization for text-to-image generative models \citep{manas2024improving}.

\noindent \textbf{Multimodal Chain-of-Thought Prompting.}
Recent works indicate that multimodal Chain-of-Thought reasoning significantly enhances the understanding capabilities of MLLMs. MM-CoT \citep{zhang2023multimodal} develops a two-stage framework where the model first learns to produce rationales using ground-truth annotations and then employs all available information to generate the final answer. DDCoT \citep{zheng2023ddcot} integrates multimodality into reasoning by initially separating the responsibilities of MLLMs into reasoning and recognition, then incorporating the visual recognition capabilities of visual models into the combined reasoning process. The work of \citet{yang2023good} introduces Question-Driven Visual Exploration (QVix), which leverages MLLMs to generate input-exploratory questions, guiding MLLMs to explore visual content more comprehensively and uncover subtle or peripheral details. CCoT \citep{mitra2024compositional} first generates a scene graph using the MLLMs and then uses that scene graph in the prompt to produce a response.
These multimodal Chain-of-Thought methods rely on guiding MLLMs through multiple steps reasoning to enhance performance. 
% Such approaches naturally outperforms single-step reasoning approaches. 
% However, through the method we propose, ~\ourmethod{}, EMLLMs can achieve or even surpass the performance of multimodal Chain-of-Thought.

%% file: section/3_method.tex
\section{Methodology}
\label{sec:method}

\begin{figure*}[t]
    \centering
    \includegraphics[width=\linewidth]{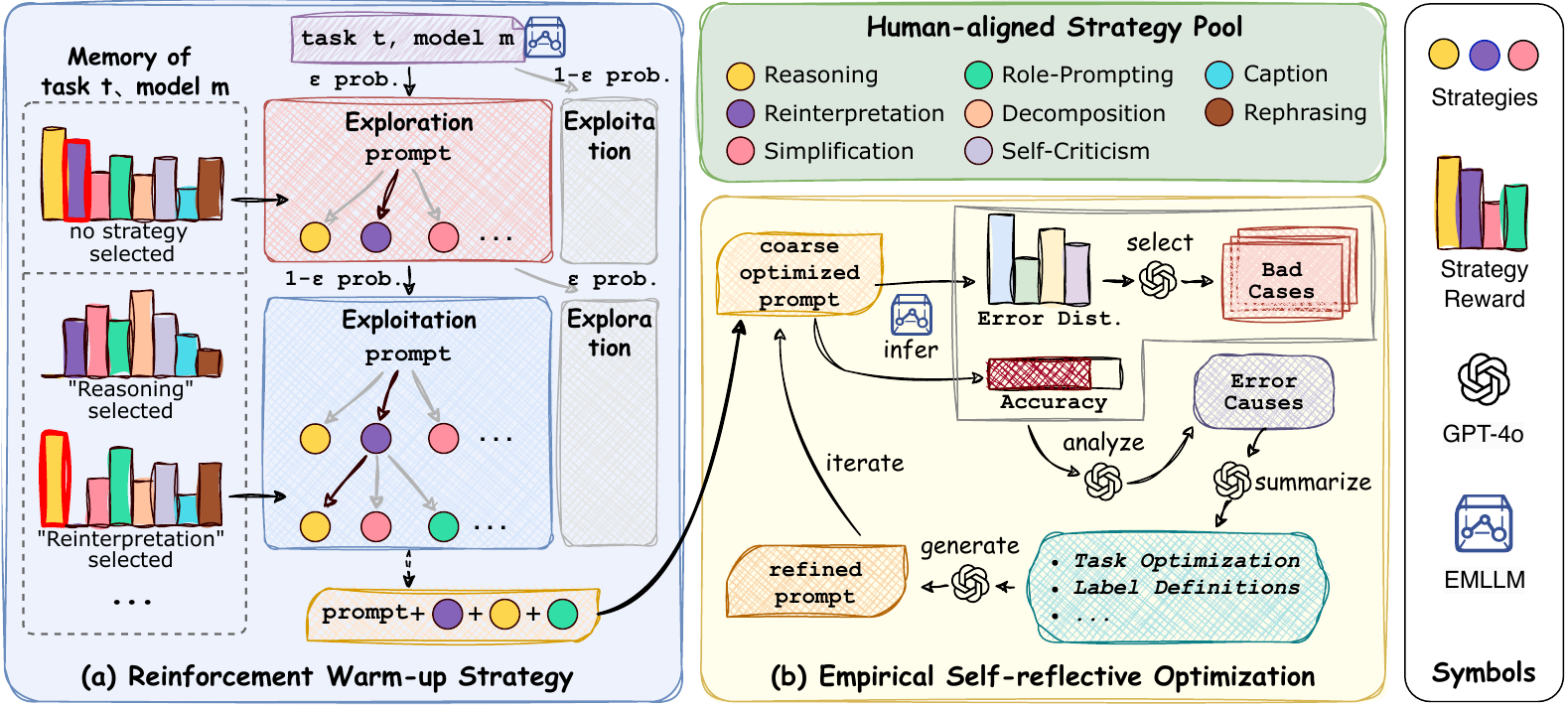}
    \caption{The architecture of \ourmethod{}. It includes Strategy Pool, \ourstagei{} and \ourstageii{}, avoiding parameter fine-tuning and requiring only a small amount of data for efficient adaptation on the device.}
    \label{fig:main}
    \vspace{-0.5cm}
\end{figure*}

This section describes the details of \ourmethod{} (Figure \ref{fig:main}). We will present each module and its optimization procedure.

\subsection{Problem Formulation and Notations}
% \subsubsection{Data}
\noindent \textbf{Data.}
% Input query $\mathcal{X}$ (\emph{e.g.}, image for vision-language models).
% Specifically, we use \( X_{\text{train}} \) to represent the private data on the device used for training, and \( X_{\text{test}} \) to represent the private data on the device after deployment. We sample high-value data from \( X_{\text{train}} \) as \( X_{\text{train}}^{\text{sub}} \subseteq X_{\text{train}} \). Correspondingly, we use \( Y_{\text{train}} \), \( Y_{\text{test}} \), and \( Y_{\text{train}}^{\text{sub}} \) to represent their corresponding ground truths.
% For a sample, \( x \) represent a sample in \( \mathcal{X} \), and \( y \) represent a sample in \( \mathcal{Y} \). 
% Given a multimodal content understanding task-$t$ that is characterized by a data distribution $\mathcal{D}_t$.
The input query of the on-device private data is represented by \( \mathcal{X} \) (\emph{e.g.}, image for vision-language models). High-value critical bad cases sampled by \ourmethod{} from \( \mathcal{X} \) is represented by \( \mathcal{X}^{\text{sub}} \subseteq \mathcal{X} \). Correspondingly, we use \( \mathcal{Y} \), and \( \mathcal{Y}^{\text{sub}} \) to represent their ground truths.
For a sample, \( x \) and \( y \) represent a sample in \( \mathcal{X} \) and \( \mathcal{Y} \) respectively. 
Given a multimodal understanding task-$t$ that is characterized by private domain data distribution $\mathcal{D}_t$. Prompt set is represented by $\mathcal{V}$, and a prompt in the $\mathcal{V}$ is represented by $v$.

\noindent \textbf{Model.}
% Given a multimodal content understanding task-$t$ that is characterized by a data distribution $\mathcal{D}_t$ and a EMLLM $f(\cdot)$, prompt optimization aims to generate a segment of text, \emph{i.e.}, the prompt $v$, where the task description is clearer and the definition is more accurately defined, which will then be applied to the black-box MLLM $f(\cdot)$ along with a input query $\mathcal{X}$ (\emph{e.g.}, image for vision-language models) such that the MLLM output $f(v, \mathcal{X})$ is able to correctly predict the ground-truth label $\mathcal{Y}$ for each $([v, \mathcal{X}], \mathcal{Y})\sim\mathcal{D}_t$.
The EMLLM adapted to \( D_t \) is represented by \( \mathcal{G}_E \), with parameters \( \Theta \). The general MLLM that assists the adaptation process of \( \mathcal{G}_E \) is represented by \( \mathcal{G}_M \). \( f_E(\cdot), f_M(\cdot) \) represents the forward propagation process of \( \mathcal{G}_E, \mathcal{G}_M \) respectively, for example, $f_E(v, x;\Theta)$ is the process of the prediction made by \( \mathcal{G}_E \) based on the prompt $v$ and the input query $x$ with parameters \( \Theta \).
% is able to correctly predict the ground-truth label $\mathcal{Y}$ for each $([v, \mathcal{X}], \mathcal{Y})\sim\mathcal{D}_t$.

\noindent \textbf{Formula.} Our proposed \ourmethod{} avoids parameter fine-tuning; therefore, the comparison of formulas between parameter fine-tuning and \ourmethod{} is as follows:

\noindent\textit{Parameter Fine-tuning:} Optimize the parameters \( \Theta \) based on the training data $\mathcal{X}$.
\begin{equation}
% \max_{\Theta} \mathbb{E}_{([v, X_{\text{train}}], Y_{\text{train}})\sim\mathcal{D}_t} h(f(v, X_{\text{train}};\Theta), Y_{\text{train}})
\max_{\Theta} \mathbb{E}_{([v, \mathcal{X}], \mathcal{Y})\sim\mathcal{D}_t} h(f_E(v, \mathcal{X};\Theta), \mathcal{Y})
\end{equation}

\noindent\textit{\ourmethod{}}:
% This problem is then commonly framed as a black-box maximization problem over the discrete language space $v\in\mathcal{V}$:
Find the optimal \( v \) and \( \mathcal{X}^{\text{sub}} \), without parameter fine-tuning and without optimize on the whole dataset.
\begin{equation}
% \max_{v\in\mathcal{V},X_{\text{train}}^{\text{sub}} \subseteq X_{\text{train}}} \mathbb{E}_{([v, X_{\text{train}}^{\text{sub}}], Y_{\text{train}}^{\text{sub}})\sim\mathcal{D}_t} h(f(v, X_{\text{train}}^{\text{sub}};\Theta), Y_{\text{train}}^{\text{sub}})
\max_{v\in\mathcal{V},\mathcal{X}^{\text{sub}} \subseteq \mathcal{X}} \mathbb{E}_{([v, \mathcal{X}^{\text{sub}}], \mathcal{Y}^{\text{sub}})\sim\mathcal{D}_t} h(f_E(v, \mathcal{X}^{\text{sub}};\Theta), \mathcal{Y}^{\text{sub}}),
\end{equation}
% \begin{equation}
% \max_{v\in\mathcal{V}} \mathbb{E}_{([v, \mathcal{X}], \mathcal{Y})\sim\mathcal{D}_t} h(f(v, \mathcal{X};\Theta), \mathcal{Y})
% \end{equation}
% where $h(f(v, \mathcal{X}), \mathcal{Y})$ is applied to measure the alignment between the EMLLM's output $f(v, \mathcal{X})$ and the ground truth $\mathcal{Y}$.
where $h(\cdot)$ is score function applied to measure the alignment between the EMLLM's output $f_E(v, x;\Theta)$ and the ground truth $y$.

\subsection{IDEALPrompt}
% When faced with the challenge that EMLLMs struggle to handle complex private domain task requirements, prompt optimization on EMLLMs is a viable approach.

We propose \textbf{\ourmethod{}}, a tuning-free, adaptive, prompt optimization framework for EMLLMs, which consists of two progressive optimization stages. Specifically, at the outset, several prompt optimization strategies are designed using human priors, referred to as the \textit{Strategy Pool} $\mathcal{S}$, which directly guides the prompts' optimization direction. Building on this, in the first stage, \textit{\ourstagei{} (\ourstageishort{})}, a strategy tree search based on reinforcement learning is conducted to acquire prior knowledge of optimization strategies. In the second stage, \textit{\ourstageii{} (\ourstageiishort{})}, the error distribution derived from the validation inference results is analyzed to identify critical bad cases, and self-reflection is subsequently applied to refine the prompts.
% 这是一个两阶段的框架，具体来说，第一阶段\ourstagei{}利用人类先验定义了several prompt优化策略，这将直接为模型提供优化的方向。在此基础上进行基于强化学习探索和利用的策略树搜索，以获得优化策略的先验；在第二阶段\ourstageii{}先根据\ourstagei{}结果来初始化prompt，然后进行迭代优化：首先进行验证推理，基于推理结果得到错误分布，由此根据错误分布采样关键的bad cases，之后进行反思学习得到新的prompt。

During the optimization process, a highly capable general MLLM is utilized as the prompt optimizer, while EMLLMs are employed as the prompt inference model.

\subsubsection{Human-aligned Strategy Pool}
While some works have used MLLMs for text gradient-based prompt optimization, 
% (\emph{e.g.}, OPRO \citep{Yang2023LargeLM} utilizes MLLMs to generate new prompts from historical information containing previously generated prompts with their values), 
we instead leverage human-designed prompt optimization strategies constructed a priori that effectively optimize prompts. 

% The work of \citep{schulhoff2024prompt} indicates a wide range of prompt optimization strategies. 
The work of \citep{schulhoff2024prompt} indicates a wide range of general prompt optimization strategies and has demonstrated the effectiveness of these strategies across various domains.
We carefully select several representative strategies that are beneficial, as identified by human experts, to form our \textit{Strategy Pool} $\mathcal{S} = \{s_1, s_2, \ldots, s_k\}$, where $s_k$ denotes each individual strategy. Specifically, we identify the following optimization strategies: \textit{Reasoning}, \textit{Reinterpretation}, \textit{Simplification}, \textit{Role-Prompting}, \textit{Decomposition}, \textit{Self-Criticism}, \textit{Caption} and \textit{Rephrasing}. The details of the strategies are provided in Appendix~\ref{sec:strategy_detail}.

Note that the strategies in strategy pool can be freely modified, removed or expanded to flexibly adapt various task requirements. In this work, we showcase the aforementioned strategies adapting to private domain data as our strategy pool. 
% Depending on these effective strategies, prompts are automatically optimized through strategy tree search in the following first stage, achieving superior results with fewer search steps.

\subsubsection{\ourstagei{}}
Based on these human-aligned strategies within the strategy pool, \ourstagei{} (\ourstageishort{}) conducts a reinforcement learning exploration-exploitation strategy tree search across tasks and models to optimize prompts, as illustrated in Figure~\ref{fig:main} (a).

% 具体的，定义的策略节点可视为强化学习的动作空间，在任务上的评估结果（我们的多模态任务即为准确率）可视为强化学习的动作收益，动作和收益的分布在memory module中维护。during \ourstagei{}的初始阶段，系统未见过任何任务，也未进行过任何策略探索，即memory module为空，此时系统面对第一个要解决的任务时，会基于预定义的策略集合，进行策略树的搜索，同步更新动作-收益分布。面对新任务时，系统会基于memory的先验信息，应用ε-greedy策略实现探索和利用的平衡。具体的，会有ε的概率选取历史有效的TOPk策略进行探索，同时会有1-ε的概率搜索其他策略，如此实现通过更少的搜索步数，实现策略树的探索和生长。
Specifically, the strategy nodes $s_k\in \mathcal{S}$ constitute the action space of reinforcement learning, while the evaluation results of private domain task-$t$ on EMLLM-$\mathcal{G}_E$ are considered action rewards. The reward distribution of actions is maintained in the memory module $\mathcal{M}$. Note that we maintain a distinct distribution for each task type within each EMLLM, \emph{i.e.}, $\mathcal{M} = \{\mathcal{M}_{t_0, {\mathcal{G}_E}_0},\ldots,\mathcal{M}_{t_n, {\mathcal{G}_E}_m}\}$.

In the initial state of \ourstageishort{}, the framework has not encountered any private domain tasks nor performed any strategy search on any EMLLMs, \emph{i.e.}, the memory module $\mathcal{M}$ is empty. 
When the EMLLM-${\mathcal{G}_E}_0$ encounters the first private domain task-$t_0$, the framework performs a complete strategy tree search based on the human-aligned strategies within the strategy pool, simultaneously updating the action-reward distribution $\mathcal{M}_{t_0, {\mathcal{G}_E}_0}$:
\begin{equation}
\mathcal{M}_{t_0, {\mathcal{G}_E}_0} = \psi(\mathcal{S}, t_0, {\mathcal{G}_E}_0),
\end{equation}
where $\psi$ denotes the tree search operation.

For a new private domain task-$t_n$ on a new EMLLM-${\mathcal{G}_E}_m$, the framework utilizes prior information stored in the memory module $\mathcal{M}$ and applies an $\varepsilon$-greedy strategy to balance exploration and exploitation.
Specifically, the general MLLM $\mathcal{G}_M$ first selects the distribution with the highest task-model similarity to task-$t_n$ and EMLLM-${\mathcal{G}_E}_m$ from the existing memory module $\mathcal{M}$ as a reference:
\begin{equation}
\mathcal{M}_{\rm ref} = \arg\max_{\mathcal{M}_{t, {\mathcal{G}_E}} \in \mathcal{M}} \rho((t, {\mathcal{G}_E}),(t_n, {\mathcal{G}_E}_m)),
\end{equation}
where $\rho$ denotes the task-model similarity.

The distribution is then smoothed and homogenized based on task-model similarity to reduce bias:
\begin{equation}
\mathcal{M}_{t_n, {\mathcal{G}_E}_m} = \sigma(\mathcal{M}_{\text{ref}}, \rho((t, {\mathcal{G}_E}),(t_n, {\mathcal{G}_E}_m))),
\end{equation}
where $\sigma$ denotes the smooth operation.

This distribution is used as the action-reward distribution $\mathcal{M}_{t_n, {\mathcal{G}_E}_m}$ for the private domain task-$t_n$ and EMLLM-${\mathcal{G}_E}_m$. An $\varepsilon$-greedy search is then performed based on this distribution, offering a 1-$\varepsilon$ probability of selecting one of top-k historically effective strategies for exploitation, while with a $\varepsilon$ probability, the framework randomly searches other strategies for exploration:
\begin{equation}
\mathcal{S}^* =
\begin{cases} 
\kappa(\mathcal{M}_{t_n, {\mathcal{G}_E}_m}) & {\rm prob.\:1-\varepsilon}, \\
\xi(\mathcal{S} \setminus \kappa(\mathcal{M}_{t_n, {\mathcal{G}_E}_m})) & {\rm prob.\:\varepsilon},
\end{cases}
\end{equation}
where $\kappa$ denotes the selection one of top-k operation, $\xi$ denotes the random selection operation. $s^*$ represents the selected strategies.
% This approach enables the framework to explore and expand the strategy tree with fewer search steps.

% \ourstagei{}充分利用学术界和业界prompt enginnering的策略先验，相比其他搜索方法更加白盒和显式，提升了prompt优化过程可解释性。此外，designed策略节点虽然一方面可作为策略guidance以减少搜索步数，但同时也可能会由于先验限制prompt优化的能力空间上限。为此，我们的方法保证了具有一定的策略搜索方向的先验性，同时还保证策略内一定的在文本空间探索的随机性，具体的，我们对每个优化策略都保留了一定的随机性，这是通过大模型依据策略先验指导小模型优化prompt实现的。同时，在策略树搜索的过程中，我们设计的强化学习的探索-利用策略，可以有效减少search steps，可快速得到面向当前任务较优的prompt。另外，我们之所以设计为两阶段的方案，是预期warm-up在不同任务和不同模型之间具备一定的可迁移性，类似大模型预训练的过程，能够获取面向某个模型和某个任务较为通用的prompt策略组合，这在实际业务场景中的多模态多任务理解时展示出巨大的通用性优势。

\ourstageishort{} fully utilizes prior knowledge of prompt engineering strategies. Compared to other search methods, it is more transparent and explicit, enhancing the interpretability of the prompt optimization process.
% Additionally, while the designed strategy nodes can serve as guidance to reduce the search steps, they may also impose a limitation on the optimization capacity due to the constraints of prior knowledge. To address this, our approach ensures a certain degree of prior direction in the strategy search while maintaining some randomness in the exploration of the text space within each strategy. Specifically, we retain a certain level of randomness in each optimization strategy, which is achieved by using a large model to guide the smaller model in optimizing the prompt based on the strategy prior. 
We expect \ourstageishort{} to exhibit a degree of transferability across different private domain tasks and models, similar to pre-training in the MLLMs' training paradigm. This stage allows for the acquisition of a more generalizable combination of prompt strategies for private domain-specific tasks, showcasing significant advantages in multimodal, multi-task understanding in real-world business scenarios.
% Thus, \ourstagei{} has been successfully completed. It is noteworthy that \ourstagei{} can be viewed as an alternating process between the exploration phase and the exploitation phase, continuously updating the memory module $M = \{ M_T, M_{T'} \ldots \}$. The rationale for adopting the exploration-and-exploitation phase approach lies primarily in two aspects: first, by utilizing strategy tree path combinations, we can achieve performance improvements compared to a single strategy while effectively reducing search steps, thus rapidly obtaining the best prompt based on optimized strategies; second, this approach demonstrates enhanced generalizability. Our \ourstagei{} also features interpretable search characteristics, enhancing transparency. This is because we define potentially effective optimization strategies, which may exhibit varying effectiveness when applied to different models and tasks. Therefore, determining how to utilize and combine these strategies becomes a significant research question. Since we have human-designed optimization strategies, each step of the optimization process is explicit and interpretable.

\subsubsection{\ourstageii{}}

% 经过 Warm-up 阶段，我们获得了初始优化的局部最优提示（coarse optimal prompt）以及针对不同任务类型的优化先验。在此基础上，反思学习（\ourstageii{}）结合主动学习进一步优化模型性能。在反思学习过程中，我们使用多模态大语言模型（MLLM）作为提示优化器，使其能够通过自主学习筛选关键样本，从这些关键样本中学习并纠正对模糊任务或模糊标签的定义【TODO：不一定是定义模糊，也有可能是小模型不理解任务定义，像你把高中知识教给小学生，需要将表达转换和映射成小模型的世界语言分布】。详细流程见图3（case-base流程演示）。
After the \ourstageishort{}, we obtained the coarse optimized prompts as well as optimized priors for various private domain tasks and EMLLMs. Building on this foundation, \ourstageii{} (\ourstageiishort{}) further refines the prompts by selecting critical bad cases from the error distribution of evaluation results and leveraging EMLLMs' self-reflection to better adapt to private domain data.

% 具体来说，\ourstageii{}的流程为：首先，基于warm-up的最优coarse optimal prompt在验证集进行推理，判断是否达到准确率要求。若达到，则结束；若未达到，进行\ourstageii{}。其次，在进行\ourstageii{}时，会结合历史的推理结果，包含历史prompt+准确率（+bad case和错误分布），然后结合以下两点进行优化：i.global的错误分布学习：利用大模型宏观分析容易混淆的标签值；ii.local 的 case-based learning：模型自动进行关键样本筛选，原则是典型性+差异性，确定易混淆、有差异度的bad case，结合ground truth分析错误原因，【TODO：通过对典型错误样本的分析，找到当前prompt在文本空间的隐式梯度，基于这个梯度在文本空间进行梯度下降优化】优化任务描述和标签定义。如此循环，直到验证集达到准确率要求，或者达到要求的迭代次数（3次在实践中）
Specifically, the process of \ourstageiishort{} is as follows: First, the coarse optimized prompts obtained from \ourstageishort{} are used as the initial prompts for \ourstageiishort{}. 
\begin{equation}
v'_0 = v_0 + \mathcal{S}^*
\end{equation}

Building upon this foundation, \ourstageiishort{} begins iterative optimization with the EMLLM first performing evaluation on the validation set. During the optimization process, historical evaluation results of the validation set are considered, including previous prompts, accuracy rates, and error distributions. Optimization by the general MLLM $\mathcal{G}_M$ is carried out based on two key aspects:
\begin{itemize}[leftmargin=1em, itemsep=0em, parsep=0em, topsep=0em]
    \item i) Global Error Distribution Learning: the general MLLM $\mathcal{G}_M$ analyzes error evaluation results to identify and assess patterns in error distribution of private domain, uncovering task descriptions and labels that are commonly misunderstood or not well comprehended by EMLLMs. This macro-level analysis provides insights into the EMLLM's weaknesses within specific tasks, guiding further optimization.
    \item ii) Local Case-based Learning: the general MLLM $\mathcal{G}_M$ selects critical bad cases according to error distribution based on the principles of typicality and diversity to identify confusing or distinctive private domain bad cases. These cases are analyzed alongside ground truth labels to diagnose errors, refining private domain tasks and label definitions as needed. 
\end{itemize}

The optimization process can be formalized as follows:
\begin{equation}
\begin{gathered}
\mathcal{R}(v'_0) = \mathbb{E}_{([v'_0, \mathcal{X}^{\text{sub}}], \mathcal{Y}^{\text{sub}})\sim\mathcal{D}_t} h(f_E(v'_0, \mathcal{X}^{\text{sub}};\Theta), \mathcal{Y}^{\text{sub}}) \\
v_{i+1} = v_i + \nabla_{v_i} \mathcal{R}(v_i) \: {\rm for} \: i = 0, 1, \ldots
\end{gathered}
\end{equation}

This process iterates until it reaches the specified number of iterations.
\begin{equation}
\mathcal{T}(v_i) = {\rm True} \: {\rm if} \:{\rm max\_iterations} \leq i 
\end{equation}

% 由此，\ourstageii{}可以在通过warm-up Learning得到的先验的基础上，在新任务上仅进行少量的搜索步数即可获得在当前任务上表现较优的prompt，就像大模型训练中的supervised fine-tuning一样.
Thus, \ourstageiishort{} can utilize the priors obtained from \ourstageishort{} to develop fine-tuned prompts for unknown private domain tasks with minimal search steps, similar to supervised fine-tuning in the MLLMs' training paradigm.

\input{tab/main_results}

\begin{figure}[t]
    \centering
    \includegraphics[width=0.95\linewidth]{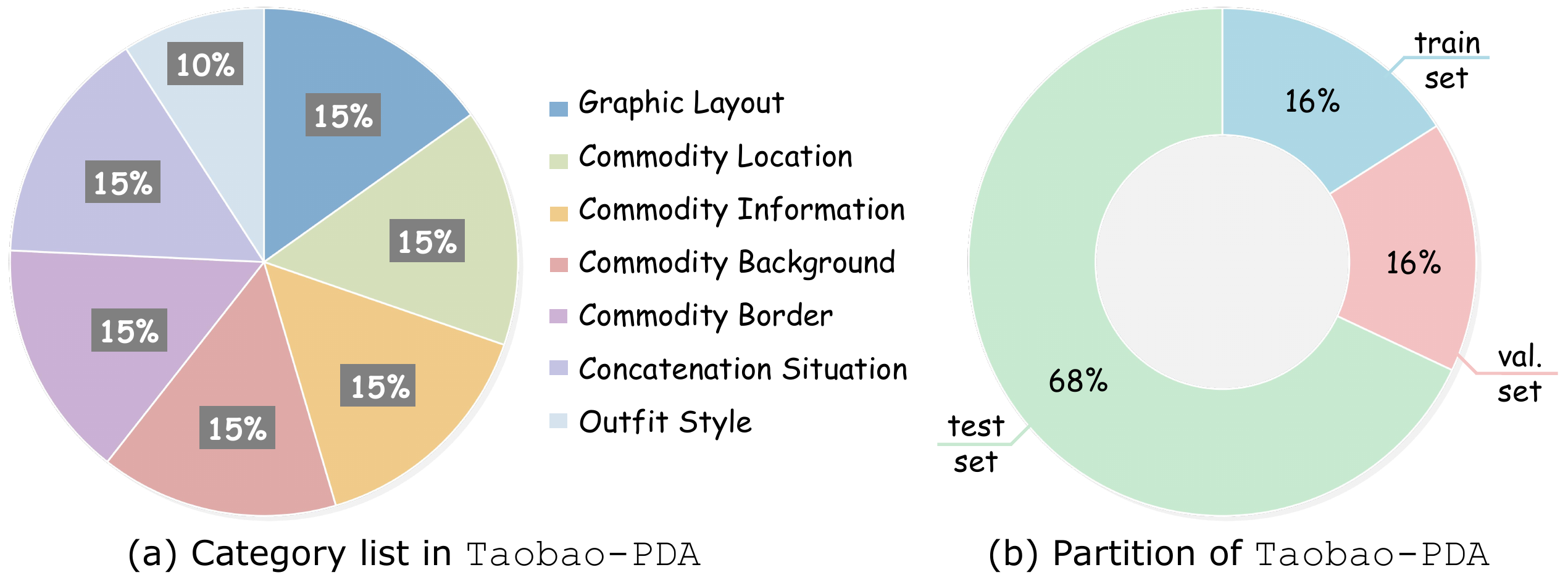}
    \caption{Data characteristics of \benchmark{}}
    \label{fig:data}
    \vspace{-0.5cm}
\end{figure}

%% file: tab/main_results.tex
\begin{table*}[t]
    \centering
    \resizebox{\textwidth}{!}{%
    \renewcommand{\arraystretch}{1.38}
    \begin{tabular}{lc||cc|cccc|c|c}
    \midrule[2pt]
    \multirow{3}{*}{\textbf{Method}} & \multirow{3}{*}{\textbf{\makecell{w/o\\ Parameter\\ Tuning}}} & \multicolumn{2}{c|}{\texttt{Image}} & \multicolumn{4}{c|}{\texttt{Commodity}} & {\texttt{Outfit}} & \multirow{3}{*}{\textbf{Average}} \\
    \Xcline{3-9}{0.2pt}
    & & \textbf{\makecell{Graphic\\ Layout}} & \textbf{\makecell{Commodity\\ Location}} & \textbf{\makecell{Commodity\\ Information}} & \textbf{\makecell{Commodity\\ Background}} & \textbf{\makecell{Commodity\\ Border}} & \textbf{\makecell{Concatenation\\ Situation}} & \textbf{\makecell{Outfit\\ Style}} \\
    \midrule[2pt]
    \textbf{Fine-Tuning} & & \cellcolor{myorange}{\textbf{38.5}} & 26.0 & \cellcolor{myred}{\textbf{69.8}} & 61.5 & \cellcolor{myorange}{\textbf{58.3}} & \cellcolor{myorange}{\textbf{88.5}} & \cellcolor{myred}{\textbf{47.1}} & \cellcolor{myorange}{\textbf{55.7}} \\
    \textbf{MM-CoT} & \faCheckCircle & 21.9 & \cellcolor{myorange}{\textbf{27.1}} & 24.0 & 42.7 & 25.0 & 38.5 & 17.6 & 28.4 \\
    \textbf{Zero-Shot} & \faCheckCircle & 16.7 & 5.2 & 14.6 & 42.7 & 54.2 & 5.2 & 14.7 & 22.5 \\
    \textbf{APE} & \faCheckCircle & 34.4 & 26.0 & 6.3 & 38.5 & 18.8 & 15.6 & 8.8 & 21.2 \\
    \textbf{OPRO} & \faCheckCircle & 20.8 & \cellcolor{myorange}{\textbf{27.1}} & 39.6 & \cellcolor{myorange}{\textbf{68.8}} & 54.2 & 32.3 & 26.5 & 38.5 \\

    \hline
    \rowcolor{blue!5} \textbf{\ourmethod{}} & \faCheckCircle & \cellcolor{myred}{\textbf{47.9}} & \cellcolor{myred}{\textbf{51.0}} & \cellcolor{myorange}{\textbf{45.8}} & \cellcolor{myred}{\textbf{69.7}} & \cellcolor{myred}{\textbf{59.4}} & \cellcolor{myred}{\textbf{89.6}} & \cellcolor{myorange}{\textbf{38.2}} & \cellcolor{myred}{\textbf{57.4}} \\
    
    \midrule[2pt]
    \end{tabular}
    }
    \caption{Performance comparison of \ourmethod{} and other baseline methods on the \benchmark{} benchmark. All methods leverage the efficient MLLM, InternVL2-2B, for inference. We color each row as the \colorbox{myred}{\textbf{best}} and \colorbox{myorange}{\textbf{second best}}. }
    \vspace{-0.5cm}
    \label{tab:main_results}
\end{table*}

%% file: section/4_experiment.tex
\section{Experiment}
\label{sec:experiment}

\subsection{Experimental Setting}
\noindent \textbf{Datasets and Tasks.}
As our primary objective is to address tasks focused on multimodal content understanding of the private domain, we collect external data from Taobao’s private domain involved multimodal content understanding and propose a benchmark, namely Taobao Private Domain Adaptation (\benchmark{}) benchmark, encompassing 7 multimodal content understanding tasks across 3 categories, where the ground truth are labeled by GPT-4o and further refined by human annotation. The detailed tasks are as follows:
\begin{itemize}[leftmargin=1em, itemsep=0em, parsep=0em, topsep=0em]
\item \texttt{Image}: Graphic Layout, Commodity Location.
\item \texttt{Commodity}: Commodity Information, Commodity Background, Commodity Border, Concatenation Situation.
\item \texttt{Outfit}: Outfit Style.
\end{itemize}

Figure~\ref{fig:data} shows the detailed characteristics of \benchmark{}, with additional details provided in Appendix~\ref{sec:data_detail}~\footnote{Note that \benchmark{}'s original language is Chinese, which is translated to English for presentation in this paper.}.

\noindent \textbf{Baselines.}
We compare \ourmethod{} with several prompt engineering baseline methods: (1) MM-CoT~\citep{zhang2023multimodal}, which generates rationales before generating the final answer; (2) APE~\citep{zhou2022large}, which generates instructions using powerful models; (3) OPRO~\citep{Yang2023LargeLM}, which leverages historical prompts, scores, and error examples to guide models generate prompts with higher scores. In addition, we compare our method with (4) Zero-Shot inference and (5) Fine-Tuning method learning on the train set of \benchmark{}. Among them, Fine-Tuning involves parameter tuning, MM-CoT employs multi-step inference, while other prompt optimization methods (\emph{i.e.}, APE and OPRO) generally utilize single-step inference without parameter tuning but prompt optimization.

\input{tab/ablation}

\noindent \textbf{Implementation Details.} 
In the experiments, we leverage GPT-4o as the general MLLM for prompt optimization and InternVL2-2B as the EMLLM for inference. 
% All experiments are performed on the proposed \benchmark{} benchmark.
We use a simple 0-1 loss as the score function $h(\cdot)$, \emph{i.e.}, $h(f_E(v, x;\Theta), y) = 1$ if $f_E(v, x;\Theta) = y$, otherwise $h(f_E(v, x;\Theta), y) = 0$.
We set the general MLLM's sampling temperature to 1 and EMLLM's sampling temperature to 0 during the whole optimization process. Details of \ourstageishort{} and \ourstageiishort{} are provided in Appendix~\ref{sec:strategy_detail} and Appendix~\ref{sec:reflect_detail} respectively.

\begin{figure*}[t]
    \centering
    \includegraphics[width=0.95\linewidth]{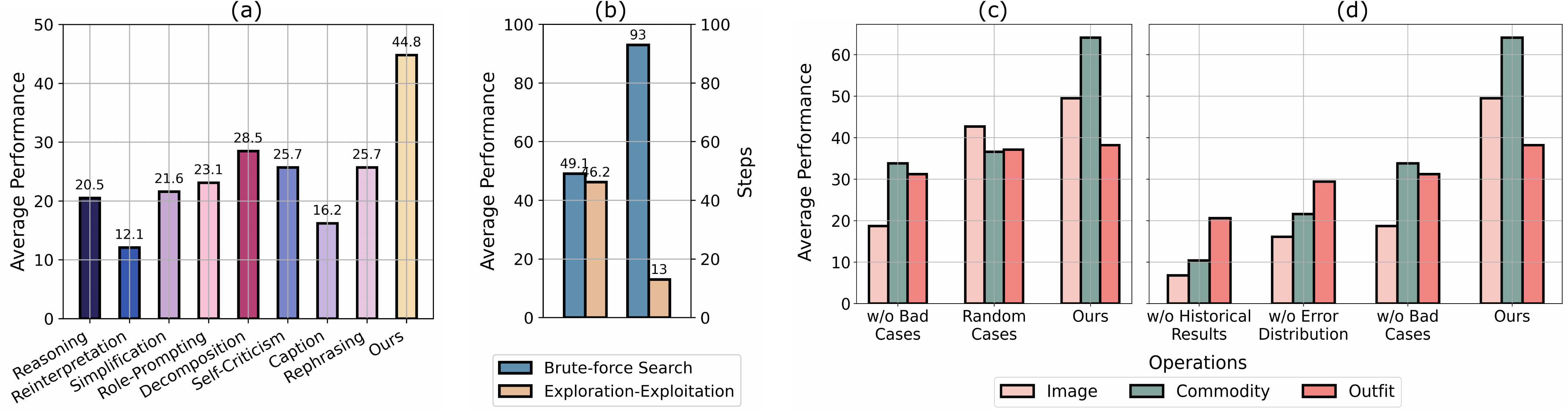}
    \caption{(a) Performance comparison between the single strategy and the exploration-exploitation strategy. (b) Performance and search steps comparison between the brute-force search and the exploration-exploitation strategy tree search. (c) and (d) Performance comparison among the absence of various components in \ourstageii{}.}
    \label{fig:ablation}
    \vspace{-0.5cm}
\end{figure*}

\subsection{Main Results}
We evaluate the performance of \ourmethod{} on the \benchmark{} benchmark compared to the baseline methods, as shown in Table~\ref{tab:main_results}. Our observations are summarized as follows: (i) \ourmethod{} achieve the highest average performance of \textbf{57.3} among all baseline methods. This result significantly surpasses that of other methods, showcasing \ourmethod{}'s superior performance across various tasks. Whether dealing with \texttt{Image}, \texttt{Commodity} or \texttt{Outfit} categories, \ourmethod{} demonstrates consistent and remarkable effectiveness. 
(ii) \ourmethod{} excel in several key tasks, achieving the best performance across 5 out of 7 tasks, \emph{i.e.}, ``Graphic Layout'' \textbf{(47.9)}, ``Commodity Location'' \textbf{(51.0)}, ``Commodity Background'' \textbf{(69.7)}, ``Commodity Border'' \textbf{(59.4)} and ``Merging Situation'' \textbf{(89.6)}. 
Particularly compared with the fine-tuning method, \ourmethod{} performs worse on only two tasks, yet slightly outperforms it in terms of overall average performance. Considering the higher costs and deployment of the fine-tuning method, this outcome is deemed acceptable.
\input{tab/compare_with_8b}
(iii) Our method consistently achieves improvements across various tasks. In contrast, other prompt optimization methods often experience either a lack of performance enhancement or even a decline. This may be attributed to the model's inability to accurately infer a reasonable optimization direction, thereby limiting their optimization capabilities.

We conduct experiments on various MLLMs for prompt optimization, experiments on public domain benchmark, human evaluation and multilingual experiments on \benchmark{}'s English version as well, see Appendix \ref{sec:exp} for details.

\subsection{Ablation Study}
To investigate the effectiveness of the two-stage framework, we conduct an ablation study comparing the complete two-stage framework with versions that omit one of the stages. Table~\ref{tab:ablation} shows that both stages of \ourmethod{} are effective, and combining both stages outperforms using each stage individually. The effectiveness of both stages avoids bias caused by introducing either stage.

\subsection{In-Depth Analysis}
\noindent \textbf{Analysis of the Strategy Optimization in RWS.}
% 在\ourstagei{}中，我们首先对比了只有单个策略优化prompt和得到的最优prompt，如图
We first compare the prompt optimized with a single strategy against the resulting coarse optimized prompt on the training set, as illustrated in Figure~\ref{fig:ablation}-(a). The ensemble strategy tree achieves a performance of 44.8, outperforming other individual strategies. Furthermore, we compare the exhaustive search of the strategy tree with the exploration-exploitation-based search, as illustrated in Figure~\ref{fig:ablation}-(b). For new tasks, exhaustive search through the strategy tree requires over 90 search steps, whereas our method needs only about 10 steps, achieving performance comparable to that of the exhaustive search.

\noindent \textbf{Effectiveness of Each Component in \ourstageiishort{}.}
% In \ourstageii{},我们对迭代优化中的各种处理进行了消融实验。具体来说，我们选择生成新prompt过程中(i)没有挑选bad cases；(ii)随机选择样本作为bad cases;(iii) 没有利用错误分布 和 (iv)没有利用历史推理结果
We conduct a performance comparison to evaluate the impact of the absence of various components within \ourstageiishort{}. Specifically, we considered the following components during the generation of new prompts: (i) without selection of bad cases; (ii) random selection of samples as bad cases; (iii) without using error distribution; and (iv) without using historical inference results. 
Figure~\ref{fig:ablation}-(c) sequentially shows the performance improvement derived from the analysis of selected bad cases, while Figure~\ref{fig:ablation}-(d) sequentially shows the performance improvement derived from analyzing conclusions based on historical evaluation results.

\begin{figure*}[t]
    \centering
    \includegraphics[width=0.95\linewidth]{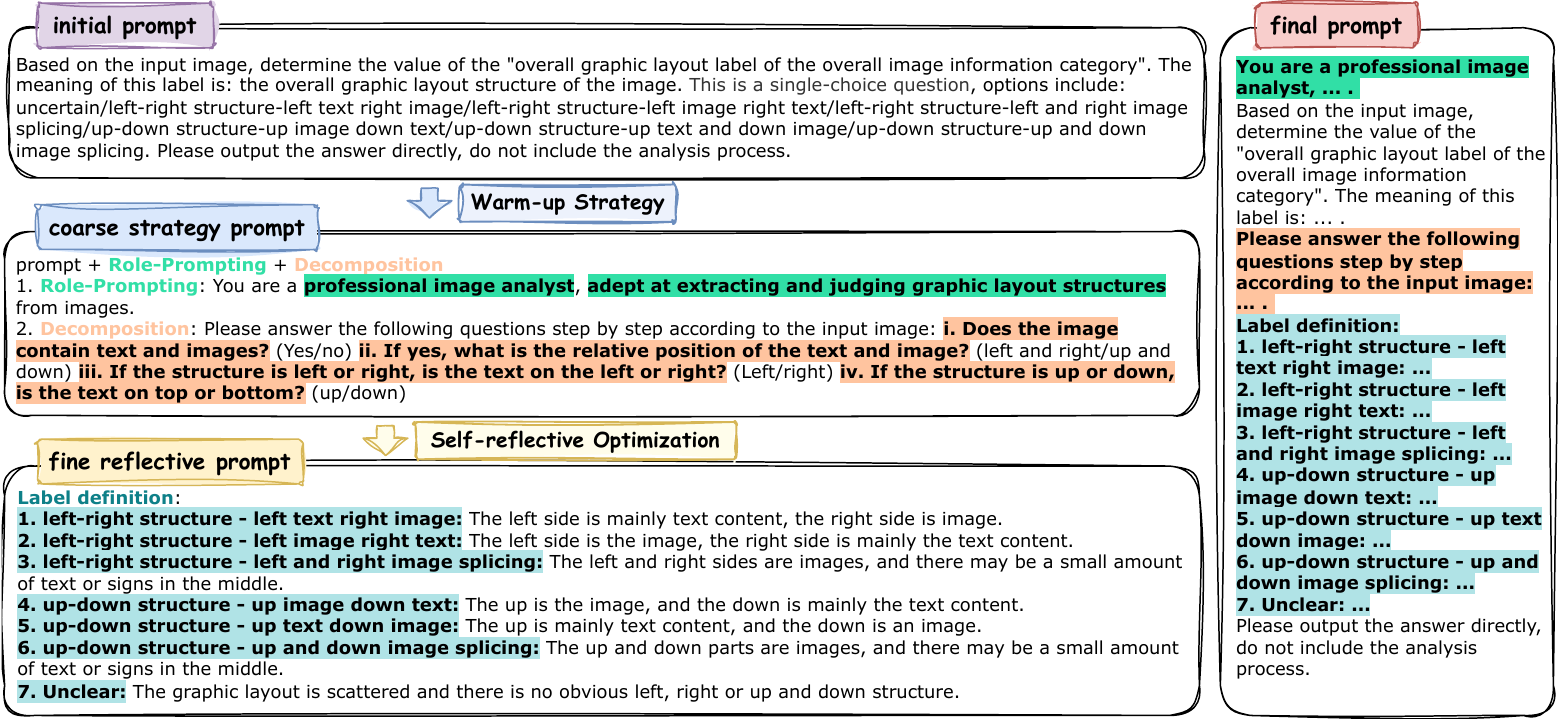}
    \caption{A case of prompt optimization using the \ourmethod{}.}
    \label{fig:case}
    \vspace{-0.2cm}
\end{figure*}

\input{tab/8b}
\input{tab/adaptability}

\noindent \textbf{Additional Baseline Comparision.}
We conduct the comparison of the average performance, inference time, and GPU memory usage between Zero-Shot on InternVL2-8B and \ourmethod{} on InternVL2-2B, as shown in Table~\ref{tab:compare_with_8b}. \ourmethod{} on InternVL2-2B achieved higher average performance than Zero-Shot on InternVL2-8B while maintaining lower inference time and GPU memory usage. In addition, we conduct the comparison of performance and average iteration steps between OPRO and \ourmethod{} on InternVL2-2B, as shown in Table~\ref{tab:steps}. \ourmethod{} achieved higher average performance than OPRO while maintaining lower iteration steps. Table~\ref{tab:8b} shows the performance comparison between Zero-Shot and \ourmethod{} on InternVL2-8B, which demonstrates the general effectiveness of our method across different models.

\noindent \textbf{Adaptability.}
Notably, our method demonstrates adaptability across both tasks and models, table~\ref{tab:adaptability} presents the adaptability of our method in task and model dimensions. Regardless of adapting to tasks or models, \ourmethod{} achieved competitive results with fewer iteration steps.

We conduct failure modes analysis, computational efficiency evaluation in Appendix~\ref{sec:analysis} and additional discussion in Appendix~\ref{sec:discussion}.

\subsection{Case Study}
Figure~\ref{fig:case} shows a case of prompt optimization using the \ourmethod{} framework. 
% The initial prompt requests the EMLLM to determine the value of the ``overall graphic layout label of the overall image information category'' based on the input image. 
After \ourstageishort{}, the optimal strategy combination, ``Role-Prompting + Decomposition'', is identified. Subsequently, \ourstageiishort{} refines the label by providing detailed definitions, ultimately producing the final optimized prompt.

%% file: tab/ablation.tex
\begin{table}[t]
\centering
\resizebox{0.9\columnwidth}{!}{% adjust width
\renewcommand{\arraystretch}{1.25}
\begin{tabular}{l||cc||c}
\midrule[2pt]
% \multirow{2}{*}{\textbf{\Large{Method}}}  
\multirow{2}{*}{\textbf{{Method}}} & 
\multicolumn{2}{c||}{\texttt{Stages}} & \multirow{2}{*}{\textbf{Average Performance}}  \\
\Xcline{2-3}{0.2pt}
& \textbf{Stage 1} & \textbf{Stage 2} & \\
\midrule[2pt]
Zero-Shot & & & 22.5  \\
w/o Self-reflection & \faCheckCircle & & 43.5  \\
w/o Warm-up Strategy &  &   \faCheckCircle  & 41.1 \\
\hline
\rowcolor{blue!5} \textbf{\ourmethod{}}   & \faCheckCircle  & \faCheckCircle  & \textbf{56.3} \\
\midrule[2pt]
\end{tabular}
}
\caption{Ablation results of involving different stages.}
\vspace{-0.5cm}
\label{tab:ablation}
\end{table}

%% file: tab/compare_with_8b.tex
% \begin{table}[t]
% \centering
% % \captionsetup{font={small,stretch=1}, labelfont={bf}}
% \caption{Comparison of average performance, inference time, and GPU memory usage between Zero-Shot results on InternVL2-8B and \ourmethod{} results on InternVL2-2B.}
% \resizebox{1\columnwidth}{!}{% adjust width
% \renewcommand{\arraystretch}{1.25}
%     \begin{tabular}{c|c|c|c|c}
%         % \Xhline{1.5pt}
%         \midrule[2pt]
%         \textbf{Method} & \textbf{MLLM} & \textbf{Average performance} & \textbf{Inference time} & \textbf{GPU memory usage} \\
%         % \hline
%         \midrule[1.5pt]
%         \textbf{Zero-Shot} & InternVL2-8B & 52.2 & \textasciitilde  & \textasciitilde 16GB \\
%         \hline
%         \textbf{\ourmethod{}} & InternVL2-2B & 56.3 & \textasciitilde  & \textasciitilde 4GB\\
%         % \Xhline{1.5pt}
%         \midrule[2pt]
%     \end{tabular}
%     }
%     \label{tab:tab4}
% \end{table}
\begin{table}[t]
\centering
\resizebox{0.95\columnwidth}{!}{% adjust width
\renewcommand{\arraystretch}{1.25}
    \begin{tabular}{l|l||c|c|c}
        % \Xhline{1.5pt}
        \midrule[2pt]
        \textbf{Method} & \textbf{\makecell{Model}} & \textbf{\makecell{Average\\Performance} } & \textbf{\makecell{Inference\\Time}} & \textbf{\makecell{GPU Memory\\Usage}} \\
        % \hline
        \midrule[1.5pt]
        \textbf{Zero-Shot} & InternVL2-8B & 52.2 & \textasciitilde 1.2$\times$ & \textasciitilde 16GB \\
        \hline
        \rowcolor{blue!5} \textbf{\ourmethod{}} & InternVL2-2B & 56.3 & \textasciitilde 1$\times$ & \textasciitilde 4GB\\
        % \Xhline{1.5pt}
        \midrule[2pt]
    \end{tabular}
    }
    \caption{Comparison of average performance, inference time, and GPU memory usage between Zero-Shot on InternVL2-8B and \ourmethod{} on InternVL2-2B.}
    \vspace{-0.5cm}
    \label{tab:compare_with_8b}
\end{table}

%% file: tab/8b.tex
% \begin{table*}
%     \centering
%     \resizebox{\textwidth}{!}{%
%     \renewcommand{\arraystretch}{1.38}
%     \begin{tabular}{l||cc|cccc|c|c}
%     \midrule[2pt]
%     \textbf{Method} & \textbf{\makecell{Graphic\\ Layout}} & \textbf{\makecell{Commodity\\ Location}} & \textbf{\makecell{Commodity\\ Information}} & \textbf{\makecell{Commodity\\ Background}} & \textbf{\makecell{Commodity\\ Border}} & \textbf{\makecell{Concatenation\\ Situation}} & \textbf{\makecell{Outfit\\ Style}} & \textbf{Average} \\
%     \midrule[2pt]
%     \textbf{Zero-Shot} & 33.3 & 52.1 & 56.3 & 38.5 & 67.7 & 76.0 & 41.2 & 52.2 \\
%     \hline
%     \rowcolor{blue!5} \textbf{\ourmethod{}} & \textbf{71.9} & \textbf{57.3} & \textbf{61.5} & \textbf{42.7} & \textbf{69.8} & \textbf{84.4} & \textbf{76.5} & \textbf{66.3} \\
    
%     \midrule[2pt]
%     \end{tabular}
%     }
%     \caption{Performance comparison between Zero-Shot and \ourmethod{} on InternVL2-8B.}
%     \vspace{-0.5cm}
%     \label{tab:8b}
% \end{table*}

\begin{table*}[t]
    \centering
    \begin{minipage}[t]{0.28\textwidth}
        \centering
        \resizebox{\columnwidth}{!}{%
        \renewcommand{\arraystretch}{1.25}
            \begin{tabular}{l||c|c}
                \midrule[2pt]
                \textbf{Method} & \textbf{ \makecell{Average \\Performance} } & \textbf{\makecell{Iteration \\Steps}} \\
                \midrule[1.5pt]
                \textbf{OPRO} & 38.5 & 50 $\pm$ 3  \\
                \hline
                \rowcolor{blue!5} \textbf{\ourmethod{}} & 56.3 & 15 $\pm$ 3 \\
                \midrule[2pt]
            \end{tabular}
        }
        \caption{Comparison between OPRO and \ourmethod{}.}
        \label{tab:steps}
    \end{minipage}\hfill
    \begin{minipage}[t]{0.705\textwidth}
        \centering
        \resizebox{\columnwidth}{!}{%
        \renewcommand{\arraystretch}{1.38}
            \begin{tabular}{l||cc|cccc|c|c}
            \midrule[2pt]
            \textbf{Method} & \textbf{\makecell{Graphic\\ Layout}} & \textbf{\makecell{Commodity\\ Location}} & \textbf{\makecell{Commodity\\ Information}} & \textbf{\makecell{Commodity\\ Background}} & \textbf{\makecell{Commodity\\ Border}} & \textbf{\makecell{Concatenation\\ Situation}} & \textbf{\makecell{Outfit\\ Style}} & \textbf{Average} \\
            \midrule[2pt]
            \textbf{Zero-Shot} & 33.3 & 52.1 & 56.3 & 38.5 & 67.7 & 76.0 & 41.2 & 52.2 \\
            \hline
            \rowcolor{blue!5} \textbf{\ourmethod{}} & \textbf{71.9} & \textbf{57.3} & \textbf{61.5} & \textbf{42.7} & \textbf{69.8} & \textbf{84.4} & \textbf{76.5} & \textbf{66.3} \\
            \midrule[2pt]
            \end{tabular}
        }
        \caption{Performance comparison between Zero-Shot and \ourmethod{} on InternVL2-8B.}
        \label{tab:8b}
    \end{minipage}
    \vspace{-0.5cm}
\end{table*}

%% file: tab/adaptability.tex
\begin{table}[t]
\centering
\resizebox{0.95\columnwidth}{!}{% adjust width
\renewcommand{\arraystretch}{1.25}
    \begin{tabular}{l|l|l||ccc|c}
        % \Xhline{1.5pt}
        \midrule[2pt]
        \multirow{2}{*}{\textbf{Type}} & \multirow{2}{*}{\textbf{EMLLM}} & \multirow{2}{*}{\textbf{Method}} & \multicolumn{3}{c|}{\textbf{Performance}} & \multirow{2}{*}{\textbf{Steps}} \\
        \Xcline{4-6}{0.25pt}
        & & & \texttt{Image} & \texttt{Commodity} & \textbf{Average} &  \\
        % \hline
        \midrule[1.5pt]
        \multirow{2}{*}{\textbf{Tasks}} & \multirow{2}{*}{InternVL2-2B} & w/o adapt & 49.0 & 49.7 & 49.5 & 93 \\
        & & \cellcolor{blue!5}{w adapt} & \cellcolor{blue!5}{49.5} & \cellcolor{blue!5}{66.1} & \cellcolor{blue!5}{60.6} & \cellcolor{blue!5}{13} \\
        \hline
        \multirow{2}{*}{\textbf{Models}} & \multirow{2}{*}{Qwen2-VL-2B} & w/o adapt & 16.1 & 62.8 & 47.2 & 93 \\
        & & \cellcolor{blue!5}{w adapt} & \cellcolor{blue!5}{17.7} & \cellcolor{blue!5}{58.6} & \cellcolor{blue!5}{45.0} & \cellcolor{blue!5}{13} \\
        % \Xhline{1.5pt}
        \midrule[2pt]
    \end{tabular}
    }
    \caption{Comparison of performance and steps between adapt from tasks/model and without adapt.}
    \vspace{-0.5cm}
    \label{tab:adaptability}
\end{table}

%% file: section/5_conclusion.tex
\section{Conclusion}
\label{sec:conclusion}
In this paper, we propose \ourmethod{}, a tuning-free, adaptive, universal prompt optimization framework consisting of two stages, which focuses on boosting EMLLMs' content understanding of private domain data.
Specifically, \ourmethod{} incorporates human experts' prior optimization strategies along with a reinforcement learning-based strategy tree search and utilizes the model's self-reflection to refine prompts.
% Specifically, \ourmethod{} 融入了人类专家先验optimization strategies 和模型的sefl-reflect.
In addition, we propose \benchmark{} benchmark to study the private domain adaptation of \ourmethod{}.
Experimental results demonstrate that \ourmethod{} outperforms existing prompt optimization approaches while significantly improving optimization efficiency.

%% file: section/6_limitation.tex
\section*{Limitations}
\label{sec:limitations}

% Compared to fine-tuning method with similar performance, \ourmethod{} will have a longer inference time when addressing tasks.
% Moreover, when training data and computational resources are abundant, \ourmethod{} tends to result in a longer training time compared to the fine-tuning method.

Our limitations and potential risks are as follows:
\begin{itemize}[leftmargin=1em]
    \item \textbf{Large Consumption of API Calls.} Our method involves extensive searching and still requires a large number of API calls (although this is much less than existing methods), which is a common issue in automatic prompt optimization.
    \item \textbf{Burden of human-aligned strategy definition.} Strategies defined by human experts are intended to provide models with human-derived optimization priors. This is because humans can usually intuitively perceive better descriptions, but optimizing them directly is difficult. However, this increases the burden of manual definitions, thereby automatic construction of external strategies will be a key area for our future research.
\end{itemize}

% \paragraph{Large Consumption of API Calls.} Our method involves extensive searching and still requires a large number of API calls (although this is much less than existing methods), which is a common issue in automatic prompt optimization.

% \paragraph{Burden of human-aligned strategy definition.} 
% Strategies defined by human experts are intended to provide models with human-derived optimization priors. This is because humans can usually intuitively perceive better descriptions, but optimizing them directly is difficult. However, this increases the burden of manual definitions, thereby automatic construction of external strategies will be a key area for our future research.

\section*{Ethics Statement}
In the process of collecting \benchmark{}, we obtained TaoBao's permission and carried out desensitization processing. In addition, racial discrimination and gender discrimination were eliminated. Especially for the data of the \texttt{outfit} category involving people, we only consider some clothes with distinct characteristics, regardless of gender or race.

%% file: section/X_suppl.tex
\section{Appendix}
\label{sec:appendix}

\begin{figure*}[b!]
    \centering
    \includegraphics[width=\linewidth]{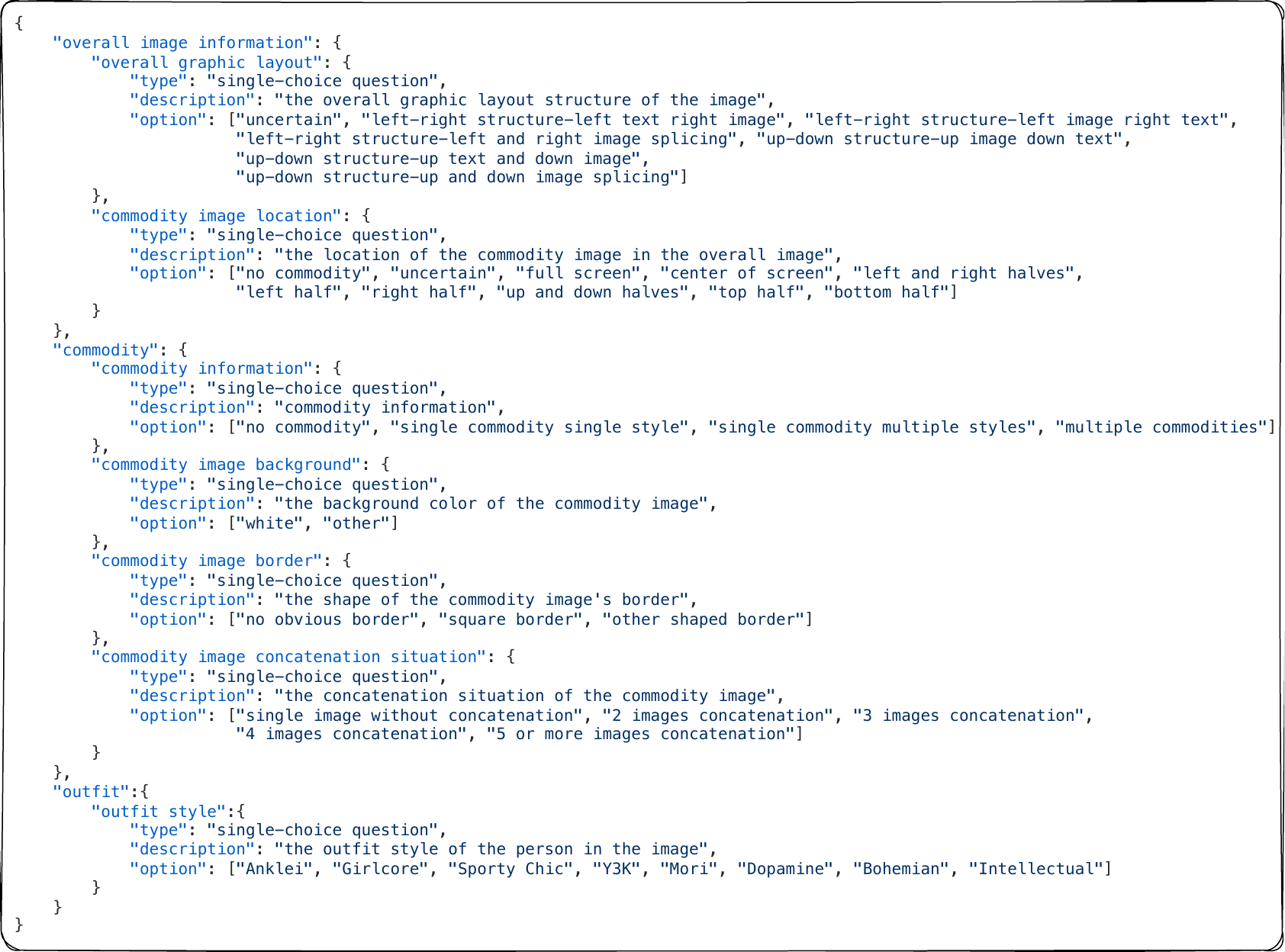}
    \caption{Tasks definition.}
    \label{fig:task}
    \vspace{-0.5cm}
\end{figure*}
\input{tab/init_prompt}

\input{tab/strategy_prompt}

\input{tab/reflect_prompt1}

\input{tab/reflect_prompt2}

\subsection{\benchmark{} Data Detail}
\label{sec:data_detail}
\subsubsection{Private Domain Data}
Private domain data refers to data characterized by highly specialized domain knowledge, which often results in models trained on public domain data performing poorly on private domain data. For instance, in our work, we focus on commodity images within the e-commerce context. The data distribution in public data typically centers on image captions/descriptions, visual question answering (VQA), grounding, and similar tasks. In contrast, the data distribution in e-commerce domain data tends to focus on the distribution and types of commoditys in the image, commodity-background boundaries, and other specific features. These differences in data distribution result in the inadequate performance of models trained on public domain data when applied to private domain data, as we demonstrate in Tbale~\ref{tab:main_results}, zero-shot performance.

In addition, while we are currently focused on the e-commerce domain, e-commerce itself encompasses a wide array of technical fields (such as image recognition, video understanding, and text analysis) and various subfields (such as apparel, fast-moving consumer goods, electronics, and healthcare). 
Therefore, we have collected data from additional modalities (such as video) and different e-commerce subfields (such as apparel and electronics).
\subsubsection{Task Definition}
Figure~\ref{fig:task} illustrates the detailed definition of tasks. Each task can be viewed as a visual question answering task in a distinct private domain of understanding. 
\subsubsection{Initial Prompt}
Our zero-shot initial prompts of various tasks are shown in Table~\ref{tab:init_prompt}.

\subsection{Strategy Detail}
\label{sec:strategy_detail}
In \ourstageishort{}, part of strategies use the \{initial prompt + strategy prompt\} as the strategy-based prompt, while other utilize GPT-4 to generate the strategy-based prompt. The top-k value is set to 3. $\varepsilon$ is set to 0.3, meaning there is 0.7 probability of exploitation and 0.3 probability of exploration. Task-model similarity is assessed by GPT-4.
\subsubsection{Strategy Definition}
\label{sec:strategy_def}
Detailed definition of strategies are followed:
\begin{itemize}[leftmargin=1em, itemsep=0em, parsep=0em, topsep=0em]
\item \textbf{Reasoning} represents that the prompt requires MLLMs to explicitly articulate their logical process and rationale within the output, enabling a deeper understanding of the question's essence and ensuring the response's validity.  
\item \textbf{Reinterpretation} represents that the prompt requires MLLMs to re-read and reinterpret the question before answering, ensuring comprehension is both accurate and contextually appropriate.  
\item \textbf{Simplification} represents that the prompt requires MLLMs to remove irrelevant information, ensuring that the content is concise and directly focused on the question.  
\item \textbf{Role-Prompting} represents that the prompt assigns a specific role to MLLMs, guiding their approach to generating responses and framing the context accordingly.  
\item \textbf{Decomposition} represents that the prompt requires MLLMs to decompose complex questions into simpler sub-instructions, addressing each sub-question individually for clarity and precision.  
\item \textbf{Self-Criticism} represents that the prompt requires MLLMs to critically reflect on their responses, identifying and addressing potential weaknesses or errors.  
\item \textbf{Caption} represents that the prompt requires MLLMs to provide a detailed description or caption of an image before generating an answer, enhancing contextual understanding and relevance.  
\item \textbf{Rephrasing} represents that the prompt requires MLLMs to thoroughly understand the task, rephrase it based on their comprehension, and outline a detailed plan for its completion.  
\end{itemize}

The above strategies are validated general prompt optimization strategies across various domains, which enhances the generalizability of our method and prevents potential overfitting to specific domains or tasks.

\subsubsection{Strategy Prompt}

Our specific prompts to add strategies by general MLLMs are shown in Table~\ref{tab:strategy_prompt}.

\subsection{Self-reflective Detail}
The number of critical bad cases is set to 5.
\label{sec:reflect_detail}
% 我们
\subsubsection{Self-reflective Prompt}
Table~\ref{tab:reflect_prompt1} and Table~\ref{tab:reflect_prompt2} shows the error analysis prompt and error summary prompt in \ourstageii{}, respectively.

\subsection{Additional Experimental Results}
\label{sec:exp}

\input{tab/other_optimizer}

\paragraph{Experiments on Various MLLMs for Optimization.}
In Table~\ref{tab:main_results}, we uses GPT-4o for prompt optimization. We supplemented closed-source models (Qwen-VL-Max, Gemini-2.0-Flash) with much lower prices than GPT-4o and a powerful open-source model (Qwen2-VL-72B-Instruct) to guide prompt optimization for experiments, as illustrated in Table~\ref{tab:other_optimizer}.

The cost of Qwen-VL-Max is approximately one-sixth of that of GPT-4o, and the cost of Gemini-2.0-Flash is about half that of GPT-4o, yet both achieve high performance. Qwen2-VL-72B-Instruct is a free and open-source model, with performance slightly lower than that of Qwen-VL-Max and Gemini-2.0-Flash.

Our method allows for the flexible replacement of the model that guides prompt optimization, enabling users to customize their selection.

\input{tab/bbh}

\paragraph{Experiments on Public Domain Benchmark.}
To evaluate the effectiveness and universality of \ourmethod{}, we conducted experiments on public domain benchmark, comparing with zero-shot and OPRO. BBH is a public domain benchmark which is often used for verifying prompt optimization; therefore, we selected part of BBH tasks for evaluation on InternVL2-2B. Following OPRO, we select 20\% of the data for each task as the training set, with the remaining 80\% used as test set. Consequently, we present accuracies in the format of ``training / test / overall (training + test)'', as illustrated in Table~\ref{tab:bbh}. The results indicate that \ourmethod{} still achieves better performance on public domain benchmark, demonstrating its effectiveness not only on private domain-specific tasks.

\begin{figure}
    \centering
    \includegraphics[width=\linewidth]{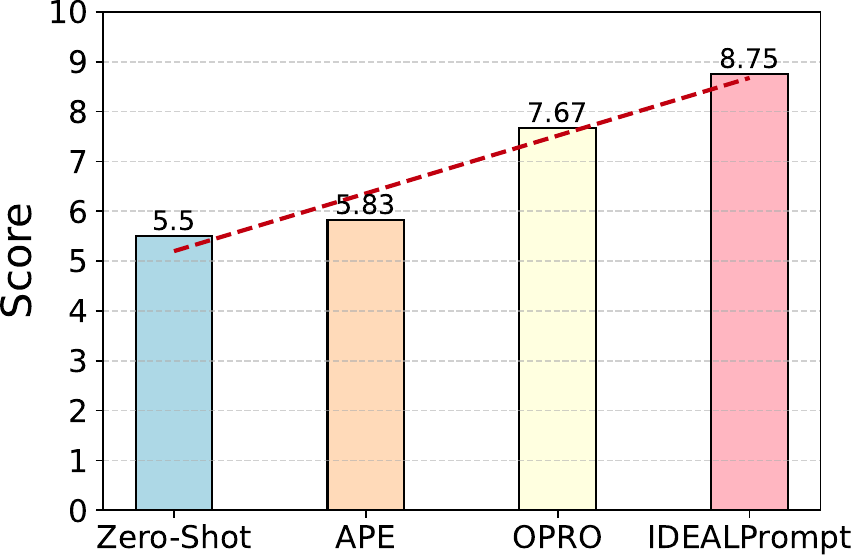}
    \caption{Comparison of human evaluation scores.}
    \label{fig:human}
    \vspace{-0.5cm}
\end{figure}

\paragraph{Human Evaluation.}
We further conduct a human evaluation on \benchmark{}. Specifically, we recruit 6 well-educated people to rank the randomly shuffled prompts optimized by APE, OPRO and \ourmethod{}. The scores range from 1 to 10 (10 means best) and are allowed to be equal for comparable instances. As shown in Figure~\ref{fig:human}. \ourmethod{} demonstrates best performance in human evaluation as well.

\input{tab/multilingual}

\paragraph{Multilingual Experiments.}
In addition, we conducted multilingual experiments to demonstrate the effectiveness of our method across multiple languages. Specifically, we constructed a batch of English prompt data and conducted relevant experiments based on the current \benchmark{} dataset, as illustrated in Table~\ref{tab:multilingual}. The results indicate that our method remains effective in English as well.

\subsection{Additional In-Depth Analysis}
\label{sec:analysis}
\subsubsection{Failure Modes Analysis}
Taking the Commodity Background task as an example, this task requires determining the background color in a commodity image, with options being ``white'' and ``others.'' Typical failure modes include: unclear task definitions, ambiguous boundaries between options, and incorrect identification of the background and its color. During self-reflection, the model's reasoning process is as follows: ``Based on the error analysis, unclear task definitions and option descriptions are the main reasons why the model fails to accurately understand the task and option meanings. Therefore, the task description in the prompt needs further simplification, and the specific meanings of the options need to be clarified. Additionally, including example images can help the model better understand the task requirements. The current prompt has partially addressed these issues, but it can be further optimized to enhance accuracy. Specifically, the prompt can more explicitly explain the distinction between `white' and other colors, and provide concrete examples to aid the model's understanding.'' Thus, it can be intuitively observed that the model can correctly analyze the failure modes of historical results.

\subsubsection{Computational Efficiency Evaluation}
The average cost of GPT-4o API calls required for a single task is approximately 0.8 RMB.
The average computation time required for a single task is approximately 15 minutes. We utilized a single A100 GPU throughout the optimization process.

\subsection{Additional Discussion}
\label{sec:discussion}
\subsubsection{Two-stage Effectiveness Discussion}
The two-stage optimization process we propose can be viewed as an optimization from coarse to fine. In the first stage, RWS, a coarse optimization based on strategies is performed. The second stage, ESO, conducts a fine optimization based on self-reflective mechanisms.

Our ablation study have validated the independent effectiveness of both stages. We posit that the two-stage optimization process not only enhances prior knowledge integration in the first-stage optimization strategies but also expedites the convergence of second-stage self-reflection mechanisms.

\subsubsection{Novelty Discussion}
\paragraph{Novelty of RWS.} Although search strategies based on reinforcement learning are widely used, our implementation in IDEALPrompt is unique in the RWS stage.
\textbf{Strategy Pool}: We have meticulously designed a strategy pool that includes various optimization strategies, which are selected and defined by human experts based on the characteristics of the private domain. This design not only enhances the flexibility of optimization but also provides richer optimization directions for the model. By defining such a strategy pool, our search efficiency has significantly improved, and we can integrate more state-of-the-art prompt strategies into the pool in the future.
\textbf{Exploration-Exploitation Strategy}: We adopt an $\varepsilon$-greedy strategy in RWS to balance exploration and exploitation. This strategy enables the model to quickly identify effective optimization strategy combinations on new tasks and models, significantly reducing the number of search steps.

\paragraph{Novelty of the Whole Framework.} Although you pointed out that some individual components (such as RL-based search and self-reflective optimization) are not entirely novel from a technical perspective, we believe that the overall framework of IDEALPrompt is distinctly innovative. By integrating the Reinforcement Warm-up Strategy (RWS) and Empirical Self-reflective Optimization (ESO), IDEALPrompt efficiently adapts to private domain data. This two-stage optimization strategy has not yet been implemented similarly in existing literature, especially in terms of achieving rapid adaptation to private data distributions without parameter fine-tuning and with only a small amount of data. Our ablation study demonstrate that both stages are highly effectiv

%% file: tab/init_prompt.tex
\begin{table*}
    \centering
    \resizebox{\textwidth}{!}{%
    \renewcommand{\arraystretch}{1.38}
    \begin{tabular}{c|c}
    \midrule[2pt]
    \textbf{Task} & \textbf{Initial Prompt} \\
    \midrule[2pt]
    \makecell{Graphic\\Layout} & \makecell[l]{\parbox[t]{18cm}{Based on the input image, determine the value of the "overall graphic layout label of the overall image information category". The meaning of this label is: the overall graphic layout structure of the image. This is a single-choice question, options include: uncertain/left-right structure-left text right image/left-right structure-left image right text/left-right structure-left and right image splicing/up-down structure-up image down text/up-down structure-up text and down image/up-down structure-up and down image splicing. Please output the answer directly, do not include the analysis process.}} \\
    \midrule[1pt]
    \makecell{Commodity\\Location} & \makecell[l]{\parbox[t]{18cm}{Based on the input image, determine the value of the "commodity image location label of the overall image information category". The meaning of this label is: the location of the commodity image in the overall image. This is a single-choice question, options include: no commodity/uncertain/full screen/center of screen/left and right halves/left half/right half/up and down halves/top half/bottom half. Please output the answer directly, do not include the analysis process.}} \\
    \midrule[1pt]
    \makecell{Commodity\\Information} & \makecell[l]{\parbox[t]{18cm}{Based on the input image, determine the value of the "commodity information label of the commodity category". The meaning of this label is: commodity information. This is a single-choice question, options include: no commodity/single commodity single style/single commodity multiple styles/multiple commodities. Please output the answer directly, do not include the analysis process.}} \\
    \midrule[1pt]
    \makecell{Commodity\\Background} & \makecell[l]{\parbox[t]{18cm}{Based on the input image, determine the value of the "commodity image background label of the commodity category". The meaning of this label is: the background color of the commodity image. This is a single-choice question, options include: white/other. Please output the answer directly, do not include the analysis process.}} \\
    \midrule[1pt]
    \makecell{Commodity\\Border} & \makecell[l]{\parbox[t]{18cm}{Based on the input image, determine the value of the "commodity image border label of the commodity category". The meaning of this label is: the shape of the commodity image's border. This is a single-choice question, options include: no obvious border/square border/other shaped border. Please output the answer directly, do not include the analysis process.}} \\
    \midrule[1pt]
    \makecell{Concatenation\\Situation} & \makecell[l]{\parbox[t]{18cm}{Based on the input image, determine the value of the "commodity image concatenation situation label of the commodity category". The meaning of this label is: the concatenation situation of the commodity image. This is a single-choice question, options include: single image without concatenation/2 images concatenation/3 images concatenation/4 images concatenation/5 or more images concatenation. Please output the answer directly, do not include the analysis process.}} \\
    \midrule[1pt]
    \makecell{Outfit\\Style} & \makecell[l]{\parbox[t]{18cm}{Based on the input image, determine the value of the "outfit style label of the outfit category". The meaning of this label is: the outfit style of the person in the image. This is a single-choice question, options include: Anklei/Girlcore/Sporty Chic/Y3K/Mori/Dopamine/Bohemian/Intellectual. Please output the answer directly, do not include the analysis process.}} \\
    \midrule[2pt]
    \end{tabular}
    }
    \caption{Initial prompts of tasks.}
    \vspace{-0.5cm}
    \label{tab:init_prompt}
\end{table*}

%% file: tab/strategy_prompt.tex
\begin{table*}
    \centering
    \resizebox{\textwidth}{!}{%
    \renewcommand{\arraystretch}{1.38}
    \begin{tabular}{c|c}
    \midrule[2pt]
    \textbf{Strategy} & \textbf{Prompt} \\
    \midrule[2pt]
    Reasoning & \makecell[l]{\parbox[t]{18cm}{\{\texttt{prompt}\} + Please carefully understand the question before answering and provide your thought and analysis process.}} \\
    \midrule[1pt]
    Reinterpretation & \makecell[l]{\parbox[t]{18cm}{\{\texttt{prompt}\} + Please do not rush to answer; before providing a response, reread and carefully understand my question and requirements.}} \\
    \midrule[1pt]
    Simplification & \makecell[l]{\parbox[t]{18cm}{You are now an expert prompt engineer, tasked with optimizing prompts to help smaller models accurately reason through complex problems. Focus on simplifying the expression of the problem and its requirements. \\My prompt is: \{\texttt{prompt}\}. \\Please output your optimized prompt directly without providing the analysis process.}} \\
    \midrule[1pt]
    Role-Prompting & \makecell[l]{\parbox[t]{18cm}{You are now an expert prompt engineer, tasked with optimizing prompts to help smaller models accurately reason through complex problems. Focus on incorporating role-playing (e.g., ``You are a xxx...'') as a method of optimization. \\My prompt is: \{\texttt{prompt}\}. \\Please output the optimized prompt directly without providing the analysis process.}} \\
    \midrule[1pt]
    Decomposition & \makecell[l]{\parbox[t]{18cm}{You are now an expert prompt engineer, tasked with optimizing prompts to help smaller models accurately reason through complex problems. Focus on decomposing the problem description into a combination of multiple simple and understandable questions to enhance performance through multi-step reasoning. \\My prompt is: \{\texttt{prompt}\}. \\Please output the optimized prompt directly without providing the analysis process.}} \\
    \midrule[1pt]
    Self-Criticism & \makecell[l]{\parbox[t]{18cm}{\{\texttt{prompt}\} + Please do not rush to answer, before giving the answer, carefully reflect on your answer is correct, confirm the answer is correct before output the answer.}} \\
    \midrule[1pt]
    Caption & \makecell[l]{\parbox[t]{18cm}{You are now an expert prompt engineer, tasked with optimizing prompts to help smaller models accurately reason through complex problems. Focus on having the model first provide a detailed description of the image content, and then formulate an answer based on this description. \\My prompt is: \{\texttt{prompt}\}. \\Please output the optimized prompt directly without providing the analysis process.}} \\
    \midrule[1pt]
    Rephrasing & \makecell[l]{\parbox[t]{18cm}{\{\texttt{prompt}\} + Did you understand the task above? Please summarize the tasks you need to do and show how you will execute the detailed plan for the task}} \\
    \midrule[2pt]
    \end{tabular}
    }
    \caption{Prompts of strategies.}
    \vspace{-0.5cm}
    \label{tab:strategy_prompt}
\end{table*}

%% file: tab/reflect_prompt1.tex
\begin{table*}[t]
    \centering
    \resizebox{\textwidth}{!}{%
    \renewcommand{\arraystretch}{1.38}
    \begin{tabular}{c}
    % \midrule[2pt]
    % \textbf{Prompt} \\
    \midrule[2pt]
    \makecell[l]{\parbox[t]{18cm}{You are currently an accomplished prompt engineer. Your task is to analyze potential causes of errors based on the model inference prompt and typical erroneous samples. These causes will be used to optimize the model inference prompt.\\
    The following is the prompt for model inference:\\ 
    \{\texttt{prompt}\}\\
    The strategies obtained in \ourstagei{} are:\\
    \{\texttt{strategies}\}\\
    The current error distribution of the model is as follows:\\
    \{\texttt{error\_distribution}\} \\
    Here are some representative error samples: \\
    \{\texttt{error\_cases}\}\\
    Please analyze the possible causes of errors from the following perspectives:\\
    1. Clarity of Task Definition:\\
    (1) Typically, models with a 2B parameter size have limited instruction-following capabilities. Is the task description in the prompt overly complex, insufficiently concise, or unclear, leading to the model's inability to comprehend the task\\
    (2) Are the descriptions of options unclear, causing the model to misunderstand the meaning of the options?\\
    (3) Are the boundaries between the options indistinct, causing the model to easily confuse certain options?\\
    2. Model Capability:\\
    (1) Although the task and options are clearly described, the model's capacity may be insufficient to solve the task. It might be necessary to attempt task decomposition or other methods to reduce task complexity.
    \\Based on the error cause analysis, please propose improvement methods for this prompt. Note that in the improvement suggestions regarding options, the names of the options must not be changed. Instead, identify the boundaries between option definitions to help the model clearly understand the specific meaning of each option and prevent confusion.
    \\Please follow this format to structure your output: \{"Error Causes": "", "Improvement Methods": ""\}}} \\
    \midrule[2pt]
    \end{tabular}
    }
    \caption{Error analysis prompt.}
    \vspace{-0.5cm}
    \label{tab:reflect_prompt1}
\end{table*}

%% file: tab/reflect_prompt2.tex
\begin{table*}[t]
    \centering
    \resizebox{\textwidth}{!}{%
    \renewcommand{\arraystretch}{1.38}
    \begin{tabular}{c}
    % \midrule[2pt]
    % \textbf{Prompt} \\
    \midrule[2pt]
    \makecell[l]{\parbox[t]{18cm}{You are currently an accomplished prompt engineer. Your task is to optimize the prompt used for inference based on historical records of prompt optimization, the current inference prompt, error distribution, and error cause analysis, with the aim of enhancing the inference performance of smaller models.\\The following are the prompts, accuracy rates, and error distribution information from previous inference rounds:\\\{\texttt{historical\_results}\}\\For the current inference, my prompt is: \\\{\texttt{prompt}\}\\The current analysis of error causes and directions for optimization are as follows:\\\{\texttt{error\_analysis\_results}\}\\Please provide the revised prompt directly, omitting the process of analysis.}} \\
    \midrule[2pt]
    \end{tabular}
    }
    \caption{Error summary prompt.}
    \vspace{-0.5cm}
    \label{tab:reflect_prompt2}
\end{table*}

%% file: tab/other_optimizer.tex
\begin{table*}
    \centering
    \resizebox{\textwidth}{!}{%
    \renewcommand{\arraystretch}{1.38}
    \begin{tabular}{l||cc|cccc|c}
    \midrule[2pt]
    \multirow{3}{*}{\textbf{Models}} & \multicolumn{2}{c|}{\texttt{Image}} & \multicolumn{4}{c|}{\texttt{Commodity}} & \multirow{3}{*}{\textbf{Average}} \\
    \Xcline{2-7}{0.2pt}
    & \textbf{\makecell{Graphic\\ Layout}} & \textbf{\makecell{Commodity\\ Location}} & \textbf{\makecell{Commodity\\ Information}} & \textbf{\makecell{Commodity\\ Background}} & \textbf{\makecell{Commodity\\ Border}} & \textbf{\makecell{Concatenation\\ Situation}} & \\
    \midrule[2pt]
    \textbf{InternVL2-2B (Zero-Shot)} & 16.7 & 5.2 & 14.6 & 42.7 & 54.2 & 5.2 & 23.1 \\
    \textbf{Qwen2-VL-72B-Instruct} & 31.3 & 41.7 & 35.4 & 52.1 & 55.2 & 64.6 & 46.7 \\
    \textbf{Qwen-VL-Max} & 36.5 & 45.8 & 35.4 & 66.7 & 55.2 & 67.7 & 51.2 \\
    \textbf{Gemini-2.0-Flash} & 36.5 & 41.7 & 38.5 & 61.5 & 57.3 & 88.5 & 54.0 \\
    \hline
    \rowcolor{blue!5} \textbf{GPT-4o} & {\textbf{47.9}} & {\textbf{51.0}} & {\textbf{45.8}} & {\textbf{69.7}} & {\textbf{59.4}} & {\textbf{89.6}} & \textbf{60.6} \\
    \midrule[2pt]
    \end{tabular}
    }
    \caption{Performance comparison of various closed-source and open-source models for prompt optimization.}
    \vspace{-0.5cm}
    \label{tab:other_optimizer}
\end{table*}

%% file: tab/bbh.tex
\begin{table}[t]
    \centering
    \resizebox{\columnwidth}{!}{%
    \renewcommand{\arraystretch}{1.38}
    \begin{tabular}{l||c|c}
    \midrule[2pt]
    \multirow{2}{*}{\textbf{Method}} & causal\_judgement & disambiguation\_qa \\
    & (train / test / overall) & (train / test / overall) \\
    \midrule[2pt]
    \textbf{Zero-Shot} & 54.1 / 49.3 / 50.3 &  62.0 / 63.0 / 62.8 \\
    \textbf{OPRO} & 78.4 / 64.0 / 66.8 &  78.0 / 75.0 / 75.6 \\
    \rowcolor{blue!5} \textbf{\ourmethod{}} & \textbf{83.8} / \textbf{64.7} / \textbf{68.4} & \textbf{82.0} / \textbf{77.0} / \textbf{78.0} \\
    \midrule[2pt]
    \end{tabular}
    }
    \caption{Accuracies on part of BBH tasks.}
    \vspace{-0.5cm}
    \label{tab:bbh}
\end{table}

%% file: tab/multilingual.tex
\begin{table}[t]
\centering
\resizebox{0.9\columnwidth}{!}{% adjust width
\renewcommand{\arraystretch}{1.25}
    \begin{tabular}{l|c}
        % \Xhline{1.5pt}
        \midrule[2pt]
        \textbf{Benchmark} & \textbf{Average Performance} \\
        % \hline
        \midrule[1.5pt]
        \textbf{\benchmark{} (Chinese)} & 57.4 \\
        \textbf{\benchmark{} (English)} & 56.3 \\
        \midrule[2pt]
    \end{tabular}
    }
    \caption{Comparison of average performance on \benchmark{} (Chinese) and \benchmark{} (English).}
    \vspace{-0.5cm}
    \label{tab:multilingual}
\end{table}